\documentclass[conference]{IEEEtran}
\IEEEoverridecommandlockouts
\usepackage{cite}
\usepackage{amsmath,amssymb,amsfonts}
\usepackage{algorithmic}
\usepackage{graphicx}
\usepackage{textcomp}
\usepackage{xcolor}
\usepackage{diagbox} 
\usepackage{graphicx}
\usepackage{subfigure} 
\usepackage{hyperref}

\newcommand\beftext[1]{{\color[rgb]{0.5,0.5,0.5}{BEFORE:#1}}}
\newcommand{\hao}[1]{{\color{orange}Hao:{#1}}}

\def\BibTeX{{\rm B\kern-.05em{\sc i\kern-.025em b}\kern-.08em
    T\kern-.1667em\lower.7ex\hbox{E}\kern-.125emX}}

\begin{document}

\title{AutoFS: Automated Feature Selection via Diversity-aware Interactive Reinforcement Learning
}

\author{\IEEEauthorblockN{Wei Fan\IEEEauthorrefmark{1}, Kunpeng Liu\IEEEauthorrefmark{1}, Hao Liu\IEEEauthorrefmark{2},  Pengyang Wang\IEEEauthorrefmark{1},  Yong Ge\IEEEauthorrefmark{3}, Yanjie Fu\IEEEauthorrefmark{1}}
\IEEEauthorblockA{\textit{\IEEEauthorrefmark{1} Department of Computer Science, University of Central Florida, Orlando} \\
\textit{\IEEEauthorrefmark{2} Business Intelligence Lab, Baidu Research, Beijing} \\
\textit{\IEEEauthorrefmark{3} Eller College of Management, University of Arizona, Tucson}\\
\IEEEauthorrefmark{1}\{weifan, kunpengliu, pengyang.wang\}@knights.ucf.edu, \\ 
\IEEEauthorrefmark{2}\{liuhao30\}@baidu.com, 
\IEEEauthorrefmark{3}\{yongge\}@arizona.edu,
\IEEEauthorrefmark{1}\{Yanjie.Fu\}@ucf.edu
}
\IEEEcompsocitemizethanks{\IEEEcompsocthanksitem \textbullet ~ Yanjie Fu is the contact author.}
\vspace{-6.5mm}
}

\maketitle

\begin{abstract}
In this paper, we study the problem of balancing effectiveness and efficiency in automated feature selection.
Feature selection is to find the optimal feature subset from large-scale feature space, and is a fundamental intelligence for machine learning and predictive analysis.
After exploring many feature selection methods, we observe a computational dilemma: 1) traditional feature selection methods (e.g., K-Best, decision tree based ranking, mRMR) are mostly efficient, but difficult to identify the best subset; 
2) the emerging reinforced feature selection methods automatically navigate feature space to explore the best subset, but are usually inefficient. 
Are automation and efficiency always apart from each other? Can we bridge the gap between effectiveness and efficiency under automation?
Motivated by such a computational dilemma, this study is to develop a novel feature space navigation method. 
To that end, we propose an Interactive Reinforced Feature Selection (IRFS) framework that guides agents by not just self-exploration experience, but also diverse external skilled trainers to accelerate learning for feature exploration.
Specifically, we formulate the feature selection problem into an interactive reinforcement learning framework. 
In this framework, we first model two trainers skilled at different searching strategies: (1) KBest based trainer; (2) Decision Tree based trainer. 
We then develop two strategies: (1) to identify assertive and hesitant agents to diversify agent training, and (2) to enable the two trainers to take the teaching role in different stages to fuse the experiences of the trainers and diversify teaching process.
Such a hybrid teaching strategy can help agents to learn broader knowledge, and  thereafter be more effective.
Finally, we present extensive experiments on real-world datasets to demonstrate the improved performances of our method: more efficient than existing reinforced selection and more effective than classic selection.

\end{abstract}


\section{Introduction}

In this paper, we aim to develop a new method to balance the effectiveness and efficiency in automated feature selection. 
Feature selection is to find the optimal feature subset from large feature space, which is a fundamental component for a wide range of data mining and machine learning tasks. 

\begin{figure}[h]
\centering
\includegraphics[height=3.7cm]{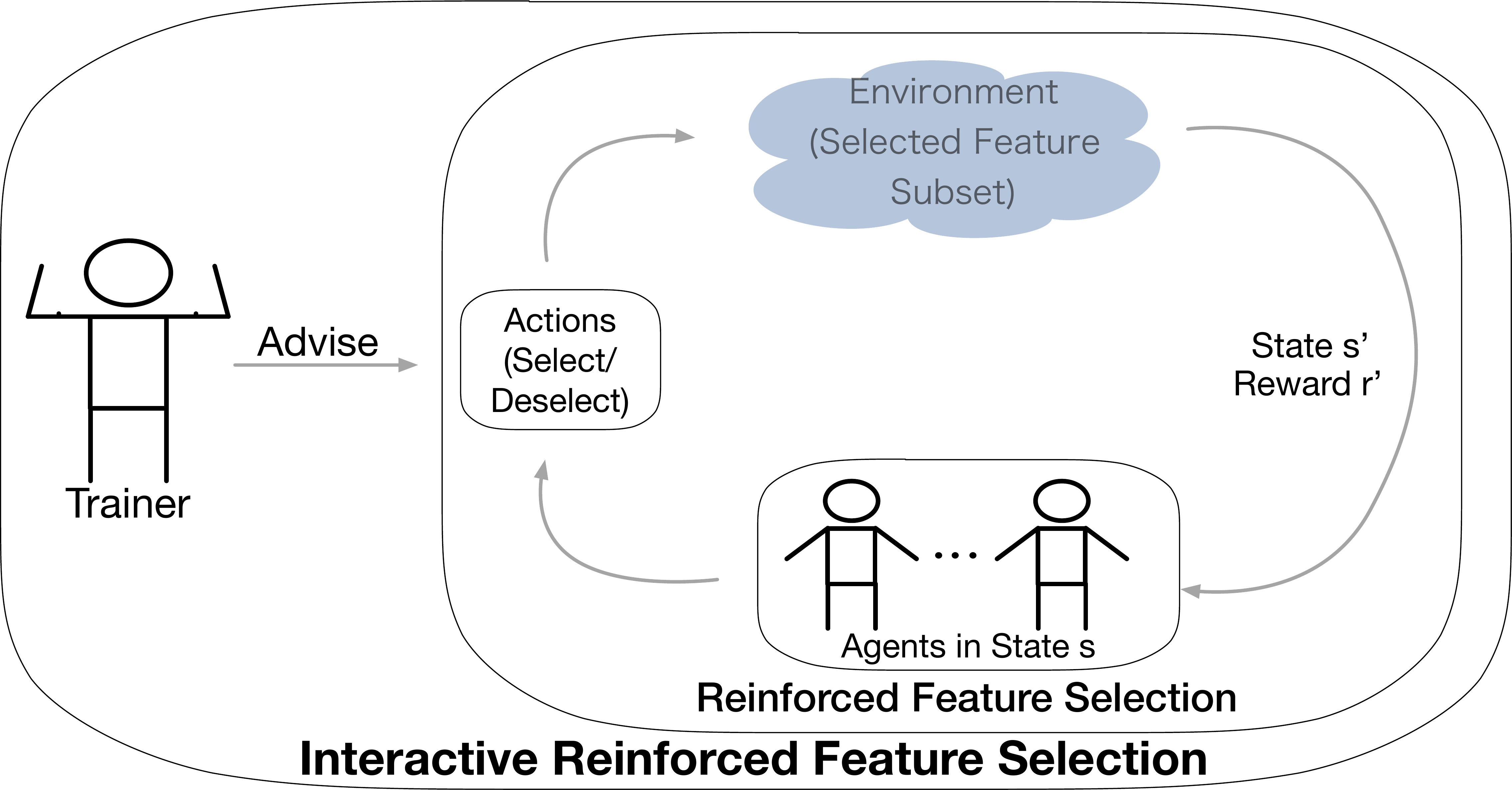}
\vspace{-3mm}
\caption{
Illustration of Interactive Reinforced Feature Selection (IRFS). Compared with basic reinforced feature selection, IRFS integrates the advice from the external teacher-like  trainer.}
\label{intro1}
\vspace{-5mm}
\end{figure}

Classic feature selection methods include: filter methods (e.g., univariate feature selection \cite{forman2003extensive}), wrapper methods (e.g., branch and bound algorithms \cite{narendra1977branch}), and  embedded methods (e.g., LASSO \cite{tibshirani1996regression}). 
Recently, the emerging reinforced feature selection methods~\cite{fard2013using,kroon2009automatic,liu2019automating} formulate feature selection  into a reinforcement learning task, in order to automate the selection process. 
Through preliminary exploration, we have observed an interesting computational dilemma in feature selection:
(1)  classic feature selection methods are mostly efficient, but difficult to identify the best subset; 
2) the emerging reinforced feature selection methods automatically navigate feature space to explore the best subset, but are usually inefficient. 
Are automation and efficiency always apart from each other? Can we bridge the gap between effectiveness and efficiency under automation to strive for a balance?
Motivated by the above dilemma, we focus on developing a new insight for addressing the effectiveness-efficiency balancing problem.
In this paper, our key insight is to integrate both self-exploration experience and external skilled trainers to simultaneously achieve both effectiveness and efficiency in feature exploration. 
To this end, two questions arise toward the goal.

First, how can we introduce external skilled trainers to improve feature selection? 
Recently, Interactive Reinforcement Learning (IRL)~\cite{cruz2018improving,suay2011effect}, as an emerging technology, has shown its superiority on accelerating the exploration of reinforcement learning through learning from external knowledge. 
Such interactive learning paradigm provides great potential for us to analogously introduce feature selection trainers to guide agents to explore and learn more efficiently.
To this end, we reformulate feature selection problem into an interactive reinforcement learning framework, which we call Interactive Reinforced Feature Selection (IRFS), as illustrated in Figure~\ref{intro1}.
We propose a multi-agent reinforcement learning component for feature selection, where an agent is a feature selector which issues select/deselect action toward its corresponding feature;
we integrate the interactive learning concept into multi-agent reinforcement learning, in order to bring in prior knowledge from external trainers (e.g., K-Best, decision tree). 
These trainers can teach agents their selection experiences by recommending actions. The agents take recommended actions to improve the efficiency of their policy functions. 
After introducing external trainers, how can we develop a new teaching strategy to enable external trainers to effectively guide agents? 
Intuitively, we directly consider a classical feature selection method as a trainer (e.g., K-Best algorithm). 
However, this initial idea has a weakness: given a static feature set, single trainer always identifies similar feature subsets; that is, the trainer only provides mostly similar or the same advice to agents every time, which jeopardizes agent learning. 
Through preliminary exploration, we have identified an interesting property of agent learning; that is, diversity matters.
Our initial insight is to diversify the set of external trainers. So, we propose a KBest based trainer and a Decision Tree based trainer, and thus both of they can teach agents to learn more about feature space. 
To improve this initial insight, we propose the second insight: to diversify the advice by selecting set of agents that will receive advice. Specifically, we randomly select a group of agents each step, which we call participated agents (features), to advise from trainers. In this way, we add randomness to improve robustness. 
Moreover, we propose the third insight:  different agents will receive personalized and tailored advice. 
Specifically, we categorize the participated agents into: assertive agents (features) and hesitant agents (features). 
The assertive agents are more confident to select their corresponding features, and  thus don't need advice from trainers. 
The hesitant agents are less confident about their decisions, and thus   need advice from trainers. 
Finally, we propose the fourth insight: to diversify the teaching process. 
Specifically, we develop Hybrid Teaching strategy to iteratively let various trainers take the teacher role at different stages, to allow agents to learn a wide range of knowledge.


In summary, in this paper, we develop a diversity-aware interactive reinforcement learning paradigm for feature selection. 
Our contributions are as follows: 
(1) We formulate the  feature selection problem into an interactive reinforcement learning framework, which we name Interactive Reinforced Feature Selection (IRFS). 
(2) We propose a teaching diversification strategy, including (a) diversifying the external trainers; 
(b) diversifying the advice that agents will receive; 
(c) diversifying the set of agents that will receive advice; 
(d) diversifying the teaching style.
(3) We conduct extensive experiments on real-world datasets to demonstrate the advantage of our methods.

\vspace{-1mm}
\section{Preliminaries}

We introduce some definitions of the reinforced feature selection; 
we present the problem formulation 
and the overview of our framework. Table \ref{method0} shows commonly used notations.

\begin{figure*}[h]
\centering
\includegraphics[width=15.6cm]{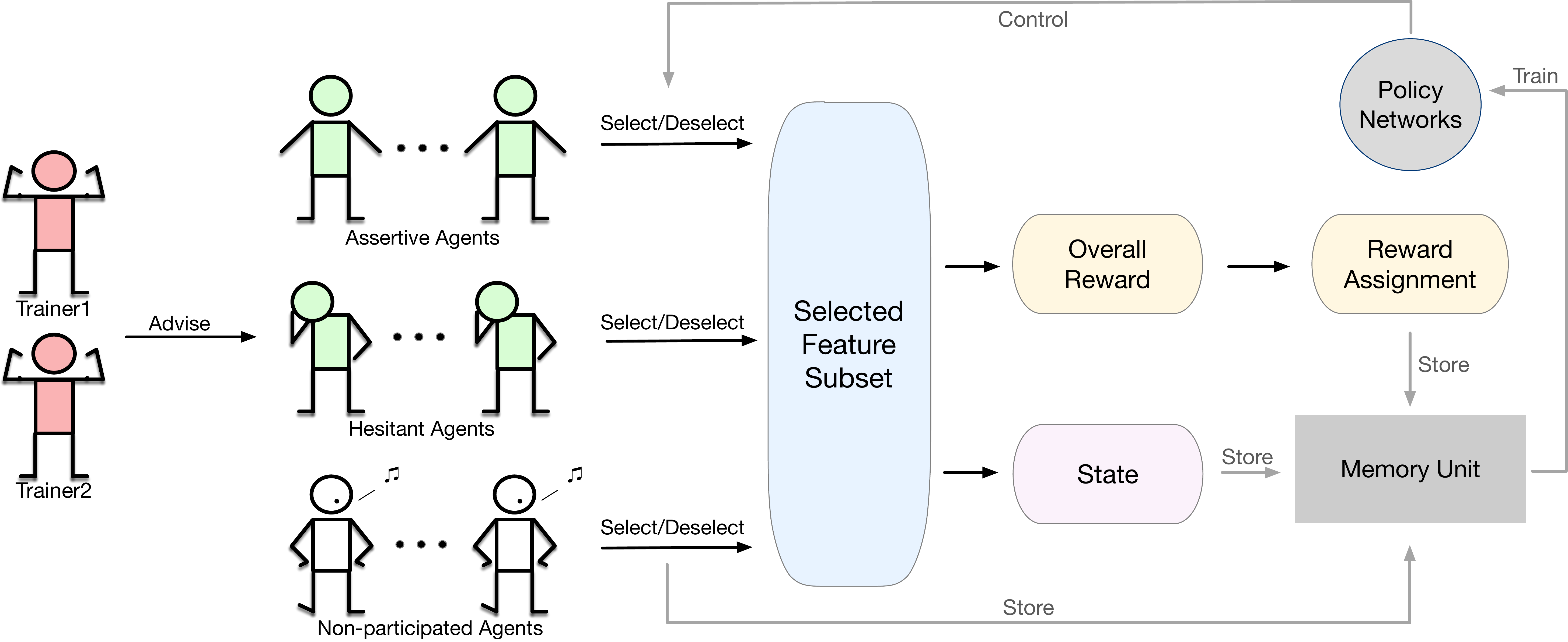}
\vspace{-3mm}
\caption{Framework Overview. Agents select/deselect their corresponding features based on their policy networks as well as the trainer's advice. Action, reward and state are stored in the memory unit for policy networks training.}
\label{overview}
\vspace{-5mm}
\end{figure*}

\subsection{Definitions and Problem Statement}

\textit{Definition 2.1.} \textbf{Agents.} Each feature $f_i$ is associated with an agent $agt_i$. After observing a state of the environment, agents use their policy networks to make decisions on the selection of their corresponding features.

\textit{Definition 2.2.} \textbf{Actions}. Multiple agents corporately make decisions to select a feature subset. For a single agent, its action space $a_i$ contains two actions, i.e., select and deselect.

\begin{table}
\vspace{-0mm}
\scriptsize
\centering
\caption{Notations.}
\vspace{-2mm}
\begin{tabular}{p{0.08\textwidth}p{0.34\textwidth}}
\hline
Notations & Definition \\
\hline
$|\cdot|$ &  The length of a set \\
$[x]$ &    The greatest integer less than or equal to x \\
$\overline{x}$ &   $1 - x$   \\
$N$ &  Number of features \\
$f_i$ &  The $i$-th feature \\
$agt_i$ &  The $i$-th agent \\
$\pi_i$ & The policy network of $i$-th agent\\
$a_i$ & The actions of $i$-th agent \\
$r_i$ &  The reward assigned to $i$-th agent \\
\hline
\end{tabular}
\label{method0}
\vspace{-4mm}
\end{table}



\textit{Definition 2.3.} \textbf{State.} The state $s$ is defined as the representation of the environment, which is the selected feature subset in each step. 

\textit{Definition 2.4.} \textbf{Reward.} The reward is to inspire the feature subspace exploration.  We measure overall reward based on the predictive accuracy in the downstream task, and equally assign the reward to the agents that select features at current step.




\textit{Definition 2.5.} \textbf{Trainer.} In the apprenticeship of reinforcement learning, the actions of agents are immature, and thus it is important to give some advice to the agents. we define the source of the advice as trainer.

\textit{Definition 2.6.} \textbf{Advice.}  Trainers give agents advice on their actions.
Multiple agents will follow the advice to take actions.

\textit{Definition 2.7.} \textbf{Problem Statement.} In this paper, we study the feature selection problem with interactive reinforcement learning. Formally, given a set of features $F =  \{f_1, f_2,  ..., f_N \}$ where $N$ is the number of features, our aim is to find an optimal feature subset $F' \subseteq F $ which is most appropriate for the downstream task. 
In this work, considering the existence of $N$ features, we create $N$ agents $ \{ agt_1, agt_2, ..., agt_N \} $ correspondingly for feature $\{f_1, f_2,  ..., f_N \}$. 
All the agents use their own policy networks $\{ \pi_1, \pi_2, ..., \pi_N \}$ to make decisions to select or deselect their corresponding features, where the actions are denoted by $\{ a_1, a_2, ..., a_N   \}$. 
For $i \in [1,N],  a_i = 1$ means agent $agt_i$ decides to select feature $f_i$; $a_i = 0$ means agent $agt_i$ decides to deselect feature $f_i$. Whenever actions are issued in a step, the selected feature subset changes, and then we can derive the state representation $s$. Finally, the reward is assigned to the certain agents based on their actions. 

\subsection{Framework Overview}
We first introduce some important components we will use in Interactive Reinforced Feature Selection (IRFS).

\subsubsection{ \textbf{Participated/Non-participated Features (Agents)}} we propose to use classical feature selection methods as trainers. However, 
the trainers only provide mostly similar or the same advice to agents every time.
To solve the problem and diversify the advice, we dynamically change the input to the trainer by selecting a set of features, which we call participated features.   We define the participated features as those features that were selected by agents in last step, e.g., if at step $t-1$ agents select $f_2, f_3,f_5$, the participated features at step $t$ are $f_2, f_3, f_5$. The corresponding agents of participated features are participated agents; other agents are non-participated agents which select/deselect their corresponding non-participated features.

\subsubsection{ \textbf{Assertive/Hesitant Features (Agents)}} we dynamically divide the participated features into \textbf{assertive features} and \textbf{hesitant features}, whose corresponding  agents  are  accordingly  called \textbf{assertive  agents} and \textbf{hesitant  agents}. Specifically, at step $t$, the participated features are divided into: the features decided to be selected by the policies are defined as assertive features, while the features decided to be deselected by the policies are defined as hesitant features. For example, at step $t$ participated features are $f_2,f_3,f_5$, and policy networks $\pi_2,\pi_3,\pi_5$ decide to select $f_2$ and deselect $f_3, f_5$. Then, assertive features are $f_2$ and assertive agents are $agt_2$; hesitant features are $f_3,f_5$ and hesitant agents are $agt_3,agt_5$.


\subsubsection{ \textbf{Initial/Advised Actions}} In each step, agents use their policy networks to firstly make action decisions, which we call \textbf{initial actions}. Then, agents take advice from external trainers and update their actions; we call the actions advised by trainers as \textbf{advised actions}. 


Figure \ref{overview} shows an overview of our framework.  There are multiple agents, each of which has its own Deep Q-Network (DQN) \cite{mnih2015human} as policy. At the beginning of each step,  each agent makes initial action decision, from which we can divide all the agents into assertive agents, hesitant agents and non-participated agents. Then trainers provide action advice to agents and agents take the advised actions.
After agents take actions, we derive a selected feature subset, whose representation is the state. 
Following \cite{liu2019automating}, we measure overall reward based on the predictive accuracy in the downstream task, and equally assign them to agents that select features at current step.
Then, a tuple of four components is stored into the memory unit, including the last state, the current state, the actions and the reward. 
In the training process, for every agent, we select mini-batches from their memory units to train their policy networks, in order to maximize the long-term reward. 

\section{Method}

We present details of Interactive Reinforced Feature Selection and diversification strategies to improve this framework.

\subsection{\textbf{Interactive Reinforced Feature Selection with KBest Based Trainer}}
\label{3.1}
\begin{figure}[h]
\centering
\includegraphics[width=8.3cm,height=8.3cm]{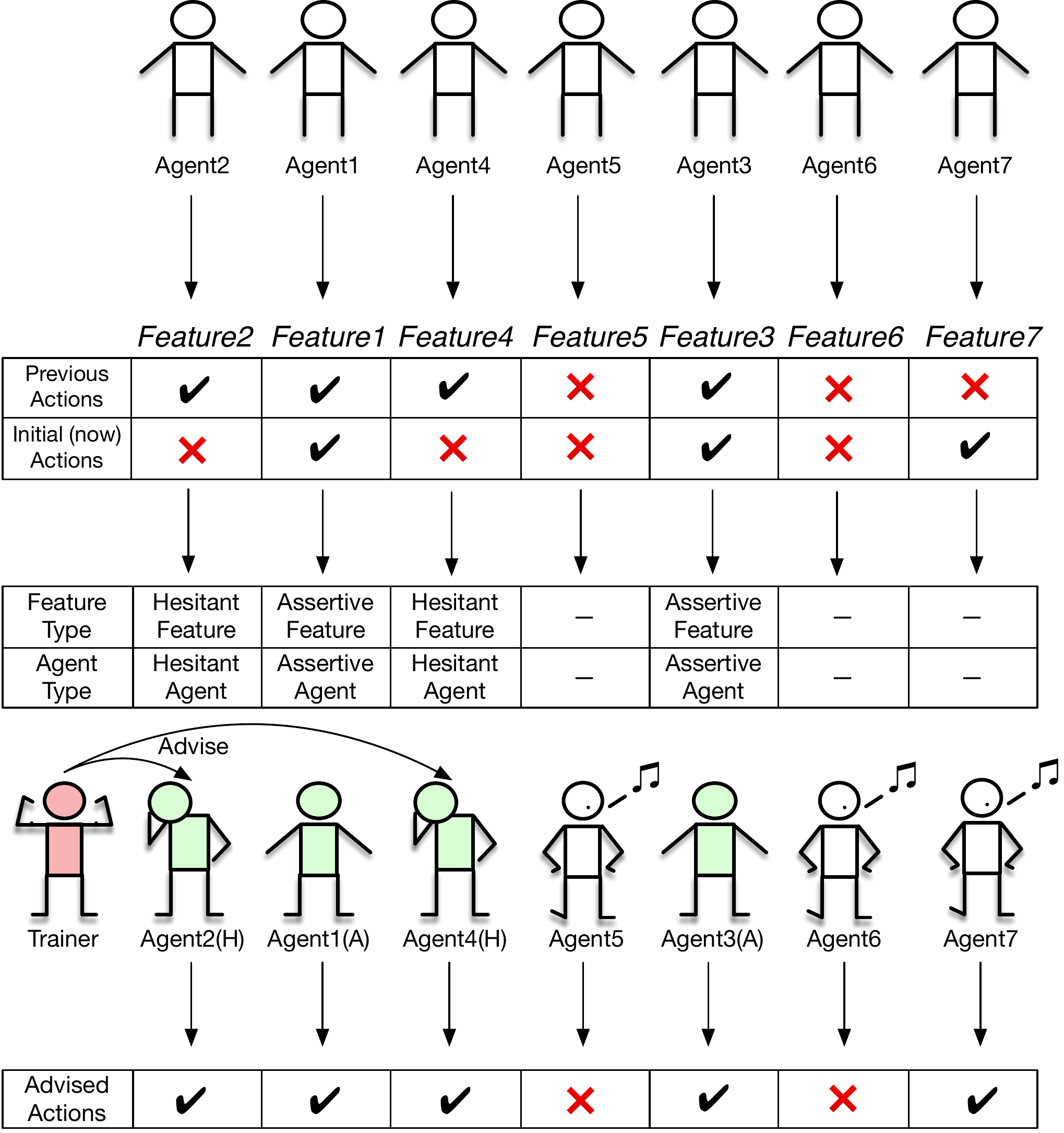}
\vspace{-3mm}
\caption{General process of trainer guiding agents. Our framework firstly identifies assertive agents (labeled by `A') and hesitant agents (labeled by `H'). Then, the trainer offers advice to hesitant agents.}
\label{method1}
\vspace{-5mm}
\end{figure}

We propose to formulate the feature selection problem into an interactive reinforcement learning framework called Interactive Reinforced Feature Selection (IRFS). Figure \ref{method1} illustrates the general process of how agents are advised by the trainer. In this formulation, we propose an advanced trainer based on a filter feature selection method, namely \textbf{KBest based trainer}. In our framework, KBest based trainer advises hesitant agents by comparing hesitant features with assertive features. We show how the trainer gives advice as follows:

\noindent\textit{\textbf{1) Identifying Assertive/Hesitant Agents}}

Given the policy networks $\{ {\pi_1}^t, {\pi_2}^t, ..., {\pi_N}^t  \}$ of multiple agents at step $t$, each agent makes an initial decision to select or deselect its corresponding feature (not actual actions to take); thus, we get an \textbf{initial} action list at step $t$, denoted by $\{ {{a_1}^t}{'}, {{a_2}^t}{'}, ..., {{a_N}^t}{'}\}$. We record actions that agents has already taken at step $t-1$, denoted by $\{ {a_1}^{t-1}, {a_2}^{t-1}, ..., {a_N}^{t-1} \}$. 
Then, we can find participated features at step $t$ by $F_p = \{\,f_i\,|\,i\in [1,N], a_i^{t-1} = 1 \}$. 
Also, we can identify assertive features as well as assertive agents. Assertive features are $F_a = \{\,f_i\,|\,i\in [1,N], f_i \in F_p \,\&\, {a_i}^{t}{'} = 1 \}$; assertive agents are $H_a = \{\,agt_i\,|\,i\in [1,N], f_i \in F_p \,\&\, {a_i}^{t}{'} = 1 \}$. 
Similarly, we can identify hesitant features and hesitant agents. i.e., hesitant features are $F_h = \{\,f_i\,|\,i\in [1,N], f_i \in F_p \,\&\, {a_i}^t{'} = 0 \}$; hesitant agents are $H_h = \{\,agt_i\,|\,i\in [1,N], f_i \in F_p \,\&\, {a_i}^{t}{'} = 0 \}$.



\noindent\textit{\textbf{2) Acquiring Advice from KBest Based Trainer}}

After identifying assertive/hesitant agents, we propose a KBest based trainer, which can advise hesitant agents to update their initial actions. 
Our perspective is: if the trainer thinks a hesitant feature is even better than half of the assertive features, its corresponding agent should change the action from deselection to selection. 

\textit{ \textbf{Step1}: (Warm-up)} we obtain the number of assertive features by $m = |F_a|$, and the number of hesitant features by $n = |F_h|$. 
We set the integer $ k = [m/2 + n]$. Then, we use K-Best algorithm to select top $k$ features in $F_p$; we denote these $k$ features by $F_{KBest}$. 

\textit{ \textbf{Step2}: (Advise)} The indices of agents which need to change actions are selected by $I_{advised} = \{ \, i \, |\, i \in [1,N], \, f_i \in F_h \,\, and \,\, f_i \in  F_{kbest} \}$. 
Finally, we can get the advised action that $agt_i$ will finally take at step $t$, denoted by $a_i^t$. Formally:
  
\begin{equation}
\centering
a_i^t=\left\{
\begin{aligned}
\overline{{a_i}^t{'}}, i \in I_{advised}  \\
{a_i}^{t}{'}, i \notin I_{advised} \\
\end{aligned}
\right.
\end{equation}

\subsection{\textbf{Interactive Reinforced Feature Selection with Decision Tree Based Trainer} }

In our IRFS framework, we propose another trainer based on a wrapper feature selection method, namely \textbf{Decision Tree based trainer}. The pipeline of IRFS with Decision Tree based trainer guiding agents could also be demonstrated in a two-phase process: (1) Identifying Assertive/Hesitant Agents; (2) Acquiring Advice from Decision Tree Based Trainer. 
The first phase is the same as phase one in Section \ref{3.1}, and the second phase is detailed as follows:



\textit{\textbf{Step1}: (Warm-up)} We denote the participated feature set as $F_p = \{f_{p_1}, f_{p_2}, ..., f_{p_t} \}$. We train a decision tree on $F_p$, and then get the feature importance for each feature, denoted by $ \{ imp_{f_{p_1}}, imp_{f_{p_2}}, ..., imp_{f_{p_t}} \}$. For hesitant features $F_h$, their importance $IMP_h = \{imp_{f_{p_j}}\;|\, f_{p_j} \in F_h  \}$; for assertive features $F_a$, their importance $IMP_a = \{imp_{f_{p_j}}\;|\, f_{p_j} \in F_a  \}$. We denote the median of $IMP_a$ by $g$.

\textit{\textbf{Step2}: (Advise)} The indices of agents which need to change actions are selected by $I_{advised} = \{\, p_j \,|\, f_{p_j} \in F_h \,\, and \,\, imp_{f_{p_j}} > g  \}$. Finally, similar to Section \ref{3.1}, we can get the advised action based on the $I_{advised}$.

\begin{table}
\small
\centering
\begin{tabular}{p{0.45\textwidth}}
\hline
Algorithm 1: IRFS with Hybrid Teacher strategy \\
\hline
\textbf{Input}: number of features: $N$, numble of exploration steps: $L$, transferring point: $T$, KBest based trainer: $trainer1(input)$, Decision Tree based trainer: $trainer2(input)$  \\

initialize state $s$, next state $s'$, memory unit $M$
 
1: \textbf{for} $t$ = 1 to $L$ \textbf{do} \\
2: \;\;\; \textbf{if} $t < T$ \textbf{do}

3: \;\;\;\;\;\;\; $\{a_1^t, a_2^t, ..., a_N^t\} \gets trainer1(input)$ \\
4: \;\;\;  \textbf{elseif} $t > T \;\& \; t < 2T $ \textbf{do}\\
5: \;\;\;\;\;\;\; $\{a_1^t, a_2^t, ..., a_N^t\} \gets trainer2(input)$\\
6: \;\;\;  \textbf{else do}\\
7: \;\;\;\;\;\;\; $\{a_1^t, a_2^t, ..., a_N^t\} \gets \{\pi_1^t,\pi_2^t, ..., \pi_N^t\}$\\

8: \;\; $\{r_1^t, r_2^t, ..., r_N^t\}$, $s' \gets$ in state $s$, agents take actions $\{a_1^t, a_2^t, ..., a_N^t\} $\\
9: \;\;\;\; $s$, $\{a_1^t, a_2^t, ..., a_N^t\}$, $s'$, $\{r_1^t, r_2^t, ..., r_N^t\}$ is stored $M$\\
10: \;\; Select mini-batch $B$ from $M$\\
11: \;\; $\{\pi_1^t,\pi_2^t, ..., \pi_N^t\}$ is trained with $B$\\
12: \;\;  $ s \gets s' $\\
13: \textbf{return} [] \\
\hline
\end{tabular}
\vspace{-4mm}
\end{table}

\subsection{\textbf{Interactive Reinforced Feature Selection with Hybrid Teaching Strategy}}

In the scenario of human being learning, teaching is commonly divided into several stages, such as elementary school, middle school and university. 
For human students, in different stages, they are always taught by different teachers who have different teaching styles and different expert knowledge. 
Inspired by the human's learning process, we propose a \textbf{Hybrid Teaching} strategy, which makes agents learn from different trainers in different periods.
Specifically, from step $0$ to step $T$, agents are  guided by one trainer.
From step $T$ to step $2T$, agents are offered advice with the help of another trainer.
Finally, after step $2T$, agents explore and learn all by themselves without the trainer. 

\begin{figure*}[htbp]
\centering
\subfigure[FC Dataset]{
\includegraphics[width=4.1cm]{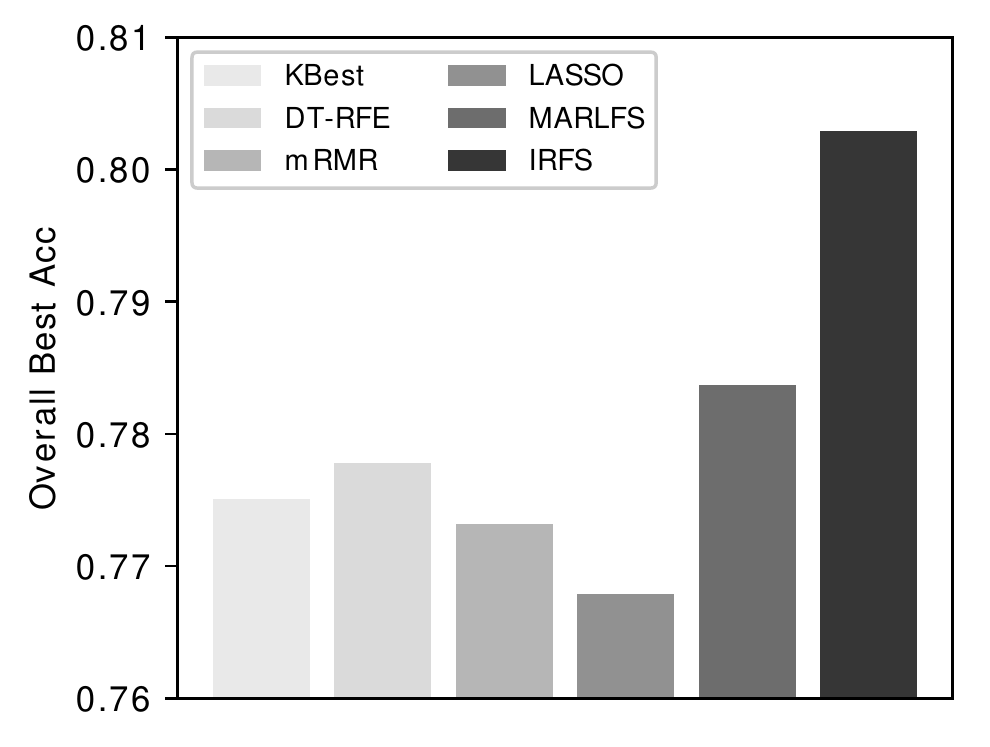}
}
\hspace{-0mm}
\subfigure[Spambase Dataset]{
\includegraphics[width=4.1cm]{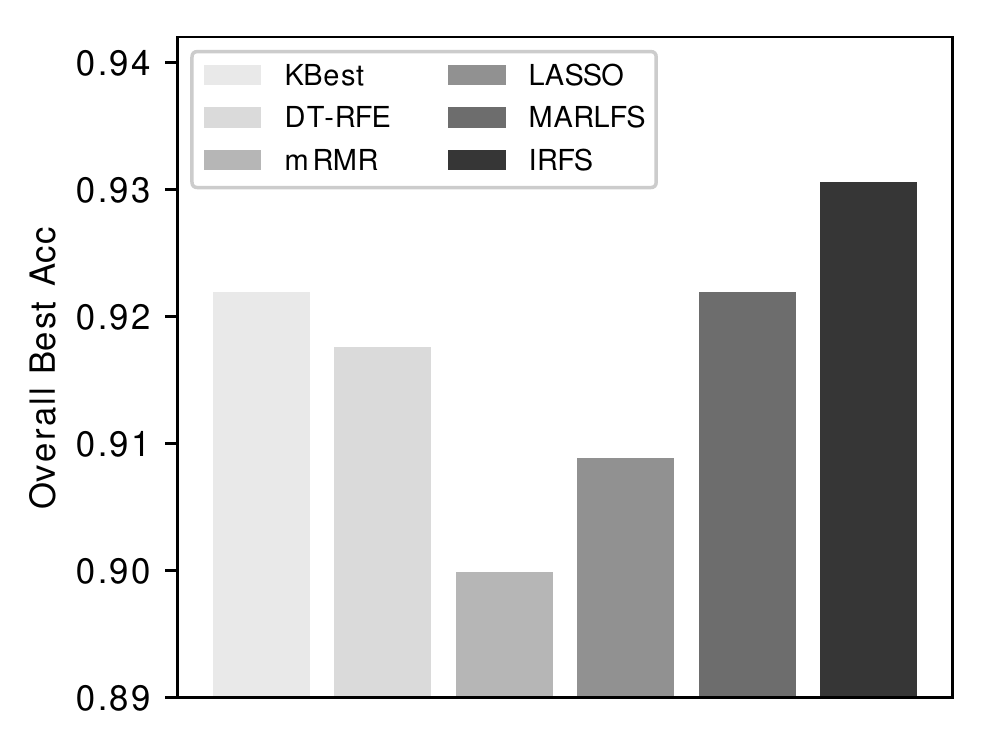}
}
\hspace{-0mm}
\subfigure[ICB Dataset]{
\includegraphics[width=4.1cm]{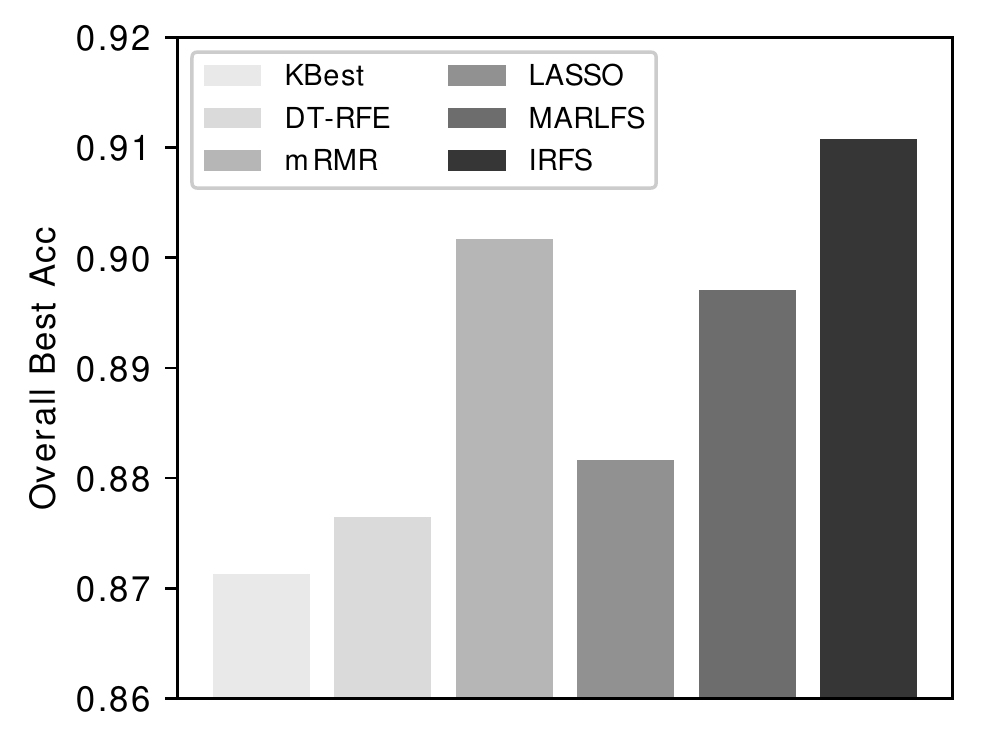}
}
\hspace{-0mm}
\subfigure[Musk Dataset]{
\includegraphics[width=4.1cm]{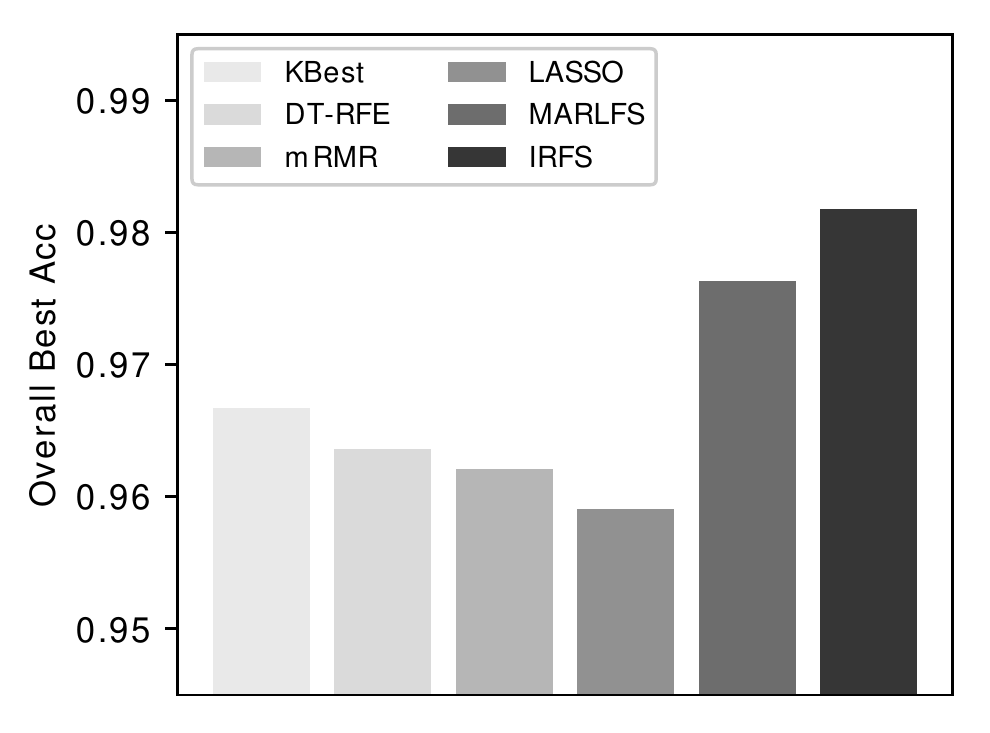}
}
\vspace{-5mm}
\caption{Overall Best Acc of different feature selection algorithms. }
\label{exp_pic1}
\vspace{-5mm}
\end{figure*}

\begin{figure*}[htbp]
\centering
\subfigure[FC Dataset]{
\includegraphics[width=4.4cm]{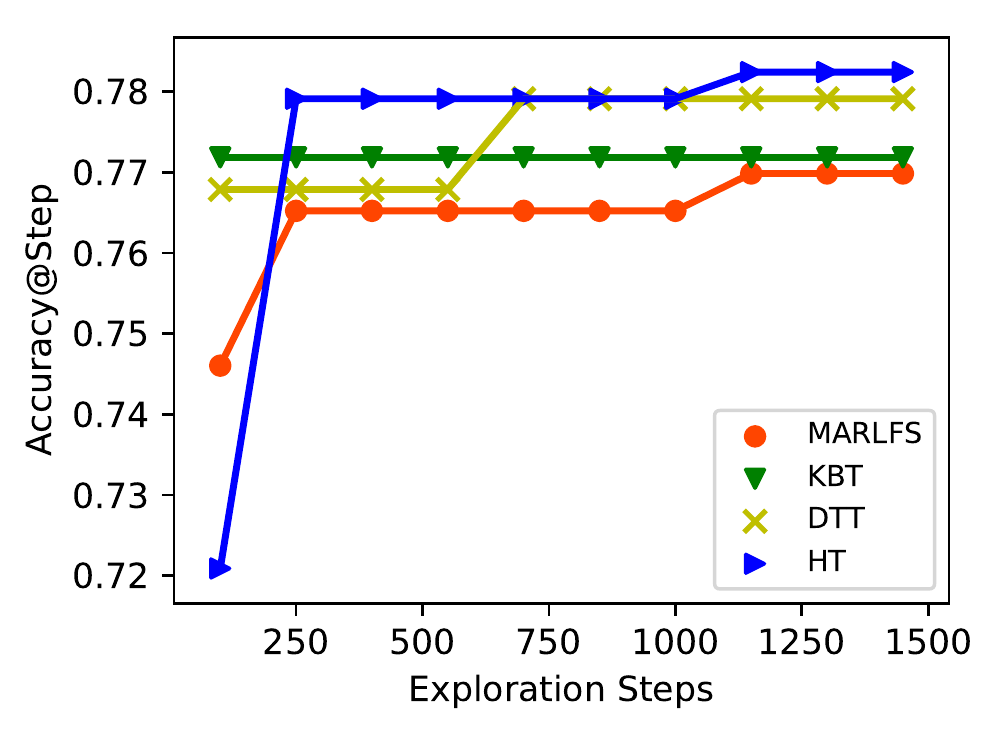}
}
\hspace{-4mm}
\subfigure[Spambase Dataset]{ 
\includegraphics[width=4.4cm]{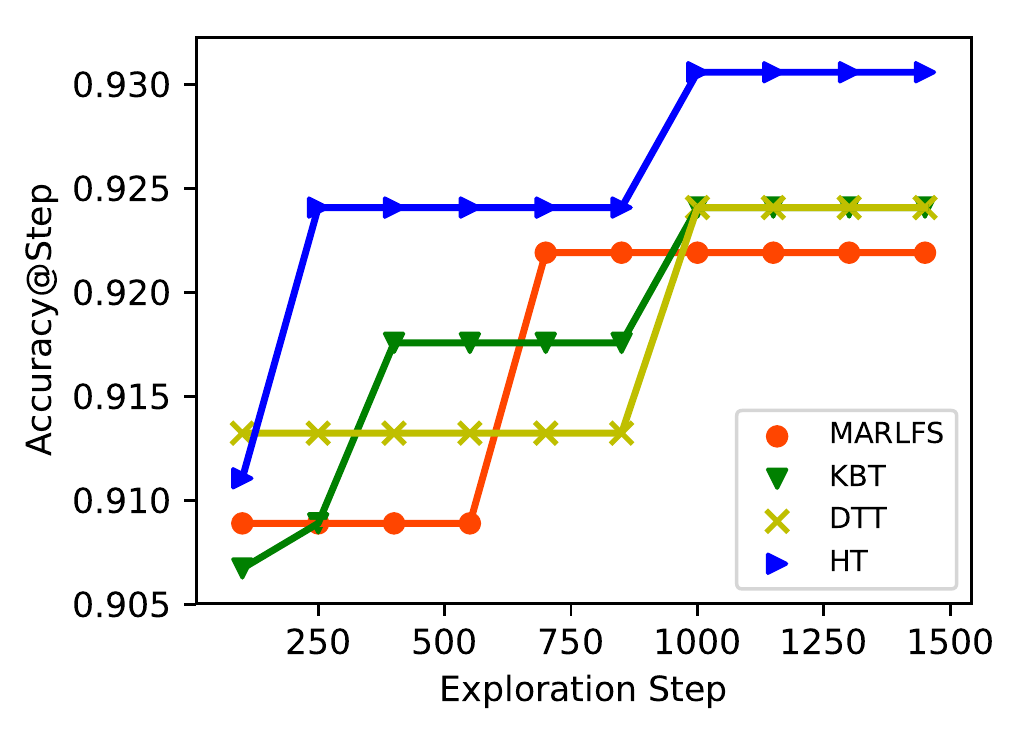}
}
\hspace{-4mm}
\subfigure[ICB Dataset]{
\includegraphics[width=4.4cm]{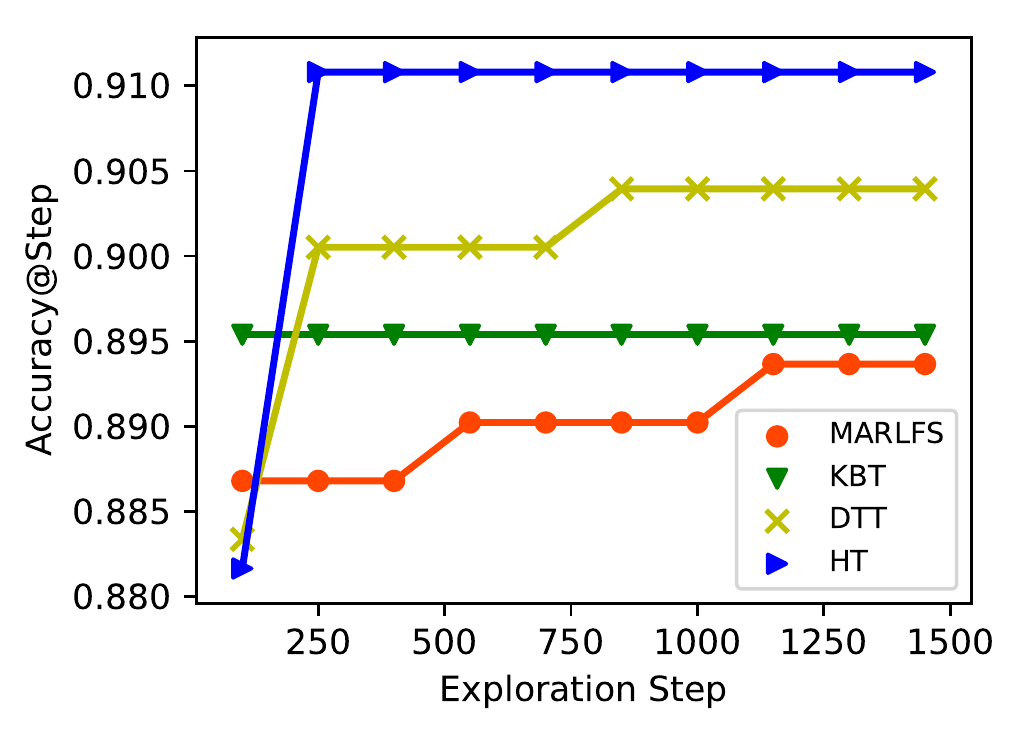}
}
\hspace{-4mm}
\subfigure[Musk Dataset]{
\includegraphics[width=4.4cm]{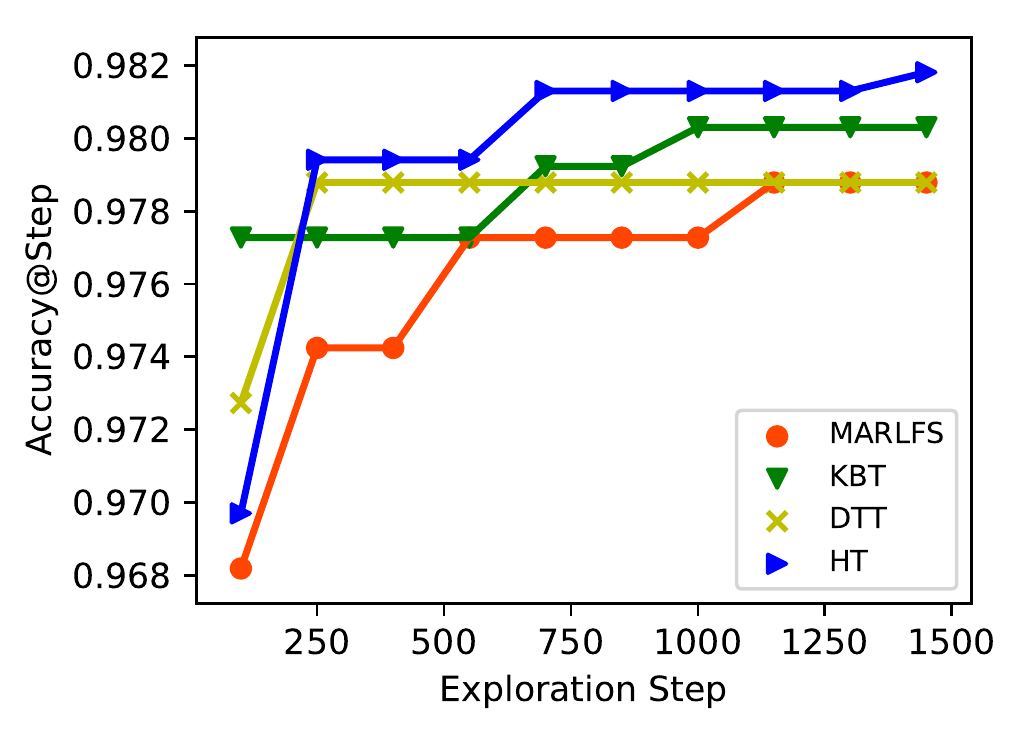}
}
\vspace{-5mm}
\caption{Exploration efficiency comparison of IRFS methods.}
\label{exp_pic2}
\vspace{-5mm}
\end{figure*}

The Hybrid Teaching strategy diversifies the teaching process and improves the exploration.
In the exploration, the randomness can lead to the similar participated features. Thus, for a single trainer, it gives similar advice when taking similar participated features as input. 
In this regard, our strategy encourages more diversified exploration by introducing different trainers with different advice though the input is similar. 
Also, advice given by only one trainer is likely to underperform, making agents suffer from unwise guidance. This strategy can provide a trade-off between advice from two trainers and thus decrease the influence of bad advice. 
Moreover, agents will lose self-study ability if they always depend on trainers' advice. Considering this, agents finally explore and learn without trainers.

\vspace{-1mm}
\section{Experimental Results}





\subsection{Data Description}
\vspace{-1mm}
\subsubsection{Forest Cover (FC) Dataset} This dataset is from Kaggle\footnote{https://www.kaggle.com/c/forest-cover-type-prediction/data} and includes 15120 samples and 54 different features.


\subsubsection{Spambase Dataset} 
This dataset includes 4601 samples and 57 different features, indicating whether a particular word or character was frequently occurring in the email \cite{Dua:2019}. 

\subsubsection{Insurance Company Benchmark (ICB) Dataset} This dataset contains information about customers, which consists of 86 variables and 5000 samples 
\cite{van2000coil,Dua:2019}. 

\subsubsection{Musk Dataset} 
It has 6598 samples and 168 features, aiming to classify the label `musk' and `non-musk' \cite{Dua:2019}.

\vspace{-1mm}
\subsection{Evaluation Metrics}
\vspace{-1mm}

\textbf{Best Acc}: Accuracy is the ratio of the number of correct predictions to the number of all predictions. 
In feature selection, an important task is to find the optimal feature subset, which has good performance in the downstream task. Considering this, we use Best Acc (BA) to stand for the performance of feature selection, which is given by $BA_{l} = max(Acc_i, Acc_{i+1}, ..., Acc_{
i+l})$, where $i$ is the beginning step and $l$ is the number of exploration steps. 

\vspace{-1mm}
\subsection{Baseline Algorithm}
\vspace{-1mm}


(1) K-Best Feature Selection. K-Best algorithm \cite{yang1997comparative} ranks features by their ranking scores with the label vector. Then, it selects the top $k$ highest scoring features. In the experiments, we set $k$ equals to half of the number of input features.

(2) Decision Tree Recursive Feature Elimination (DT-RFE). 
DT-RFE \cite{granitto2006recursive} first trains a decision tree on all the features and get the feature importance; then it recursively deselects the least important features. In the experiments, we set the selected feature number half of the feature space.

(3) mRMR. The mRMR \cite{peng2005feature} ranks features by minimizing feature’s redundancy and maximizing their relevance and selects the top-$k$ ranked features as the  result. In experiments, we set $k$ equals to half of the number of input features.

(4) LASSO. LASSO \cite{tibshirani1996regression} conducts feature selection and shrinkage via $l1$ penalty. In this method, features whose coefficients are 0 will be dropped. All the parameter settings are the same as \cite{liu2019automating}.

(5) Multi-agent Reinforcement Learning Feature Selection (MARLFS). MARLFS \cite{liu2019automating} is a basic multi-agent reinforced feature selection method, which could be seen as a variant of our proposed IRFS without any trainer. 

In the experiments, our KBest based trainer uses mutual information to select features, supported by scikit-learn. 
Our Decision Tree based teacher uses decision tree classifier with default parameters in scikit-learn. 
In state representation, following previous work \cite{liu2019automating}, we utilize graph convolutional network (GCN) \cite{kipf2016semi} to update features. 
In the experience replay \cite{lin1992self}, each agent has its memory unit. For agent $agt_i$ at step $t$, we store a tuple $\{s_i^t, r_i^t, s_i^{t+1}, a_i^t\}$ to the memory unit. 
The policy networks are set as two linear layers of 128 middle states with ReLU as activation function. In the exploration process, the discount factor $\gamma$ is set to 0.9, and we use $\epsilon$-greedy exploration with $\epsilon$ equals to 0.9.
To train the policy networks, we select mini-batches with 16 as batch size and use Adam Optimizer with a learning rate of 0.01. 
To compare fairly with baseline algorithms, we fix the downstream task as a decision tree classifier with default parameters in scikit-learn. We randomly split the data into train data ($90\%$) and test data ($10\%$). 
All the evaluations are performed on Intel E5-1680 3.40GHz CPU in a x64 machine with 128GB RAM.




\vspace{-1mm}
\subsection{Overall Performance}
\vspace{-1mm}

We compare our proposed method with baseline methods in terms of overall Best Acc on different real-world datasets. In general, Figure \ref{exp_pic1} shows our proposed Interactive Reinforced Feature Selection (IRFS) method achieves the best overall performance on four datasets. The basic reinforced feature selection method (MARLFS) has better performance than traditional feature selection methods in most cases, while our IRFS shows further improvement compared to MARLFS.

\vspace{-1mm}
\subsection{Study of Interactive Reinforced Feature Selection}
\vspace{-1mm}
We aim to study the impacts of our interactive reinforced feature selection methods on the efficiency of exploration. Our main study subjects include KBest based trainer, Decision Tree based trainer and Hybrid Teaching strategy. 
Accordingly, we consider four different methods: 
(i) \textbf{MARLFS}: The basic reinforced feature selection method, which could be seen as a variant of IRFS without any trainer.
(ii) IRFS with \textbf{KBT} (KBest based Trainer): a variant of IRFS with only KBT as the trainer. 
(iii) IRFS with \textbf{DTT} (Decision Tree based Trainer): a variant IRFS with only DTT as the trainer. 
(iv) IRFS with \textbf{HT} (Hybrid Teaching strategy): a variant of IRFS using Hybrid Teaching strategy with two proposed trainers.
Figure \ref{exp_pic2} shows the comparisons of best accuracy over exploration steps on four datasets. We can observe that both KBT and DTT could reach higher accuracy than MARLFS in few steps (1500 steps), which signifies the trainer's guidance improves the exploration efficiency by speeding up the process of finding optimal subsets. Also, IRFS with HT could find better subsets in short-term compared to MARLFS, KBT and DTT. A potential interpretation is Hybrid Teaching strategy integrates experience from two trainers and takes advantage of broader range of knowledge, leading to the further improvement.

\vspace{-1mm}
\section{Related Work}

\vspace{-1mm}
Traditional feature selection can be grouped into: 
(1) Filter methods calculate relevance scores, rank features and then select top-ranking ones (e.g. univariate feature selection \cite{forman2003extensive}). (2) Wrapper methods make use of predictors, considering the prediction performance as objective function \cite{guo2007semantic}. (e.g. branch and bound algorithms \cite{kohavi1997wrappers}). (3) Embedded methods directly incorporate feature selection as part of predictors.(e.g. LASSO \cite{tibshirani1996regression}).  
Recently, reinforcement learning \cite{sutton2018reinforcement} has been applied into different domains, such as robotics \cite{lin1991programming}, urban computing \cite{wang2020incremental}, and feature selection. 
Some reinforced feature selection studies create a single agent to make decisions \cite{ fard2013using,kroon2009automatic} but its action space is too large. Others create multi-agents to make decisions \cite{liu2019automating}. 
Reinforcement Learning \cite{sutton2018reinforcement} is a good method to address optimal decision-making problem. 
An intuitive idea to speed up the learning process is to include external advice in the apprenticeship \cite{cruz2018improving}. In Interactive Reinforcement Learning (IRL), an action is interactively encouraged by a trainer with prior knowledge \cite{knox2013teaching}. 
Using a trainer to directly advise on future actions is known as policy shaping \cite{cederborg2015policy}. 
For advice, they could come from humans and robots \cite{lin1991programming} and also from previously trained artificial agent \cite{cruz2014improving,taylor2014reinforcement}. 

\vspace{-1mm}
\section{Conclusion}

\vspace{-1mm}
In this paper, we studied the problem of balancing the effectiveness and efficiency of automated feature selection. We formulate feature selection into an interactive reinforcement learning framework, where we propose KBest based trainer and Decision Tree based trainer. To diversify the teaching process, we identify assertive and hesitant agents to give advice, and we develop a Hybrid Teaching strategy to integrate experience from two trainers. 
Finally, we present extensive experiments which illustrate the improved performance.

\section{Acknowledgment}
\vspace{-1mm}
This research was partially supported by the National Science Foundation (NSF) via the grant numbers: 1755946, I2040950, 2006889.

\bibliographystyle{IEEEtran}
\bibliography{ref}

\begin{thebibliography}{10}
\providecommand{\url}[1]{#1}
\csname url@samestyle\endcsname
\providecommand{\newblock}{\relax}
\providecommand{\bibinfo}[2]{#2}
\providecommand{\BIBentrySTDinterwordspacing}{\spaceskip=0pt\relax}
\providecommand{\BIBentryALTinterwordstretchfactor}{4}
\providecommand{\BIBentryALTinterwordspacing}{\spaceskip=\fontdimen2\font plus
\BIBentryALTinterwordstretchfactor\fontdimen3\font minus
  \fontdimen4\font\relax}
\providecommand{\BIBforeignlanguage}[2]{{%
\expandafter\ifx\csname l@#1\endcsname\relax
\typeout{** WARNING: IEEEtran.bst: No hyphenation pattern has been}%
\typeout{** loaded for the language `#1'. Using the pattern for}%
\typeout{** the default language instead.}%
\else
\language=\csname l@#1\endcsname
\fi
#2}}
\providecommand{\BIBdecl}{\relax}
\BIBdecl

\bibitem{forman2003extensive}
G.~Forman, ``An extensive empirical study of feature selection metrics for text
  classification,'' \emph{Journal of machine learning research}, vol.~3, no.
  Mar, pp. 1289--1305, 2003.

\bibitem{narendra1977branch}
P.~M. Narendra and K.~Fukunaga, ``A branch and bound algorithm for feature
  subset selection,'' \emph{IEEE Transactions on computers}, no.~9, pp.
  917--922, 1977.

\bibitem{tibshirani1996regression}
R.~Tibshirani, ``Regression shrinkage and selection via the lasso,''
  \emph{Journal of the Royal Statistical Society: Series B (Methodological)},
  vol.~58, no.~1, pp. 267--288, 1996.

\bibitem{fard2013using}
S.~M.~H. Fard, A.~Hamzeh, and S.~Hashemi, ``Using reinforcement learning to
  find an optimal set of features,'' \emph{Computers \& Mathematics with
  Applications}, vol.~66, no.~10, pp. 1892--1904, 2013.

\bibitem{kroon2009automatic}
M.~Kroon and S.~Whiteson, ``Automatic feature selection for model-based
  reinforcement learning in factored mdps,'' in \emph{2009 International
  Conference on Machine Learning and Applications}.\hskip 1em plus 0.5em minus
  0.4em\relax IEEE, 2009, pp. 324--330.

\bibitem{liu2019automating}
K.~Liu, Y.~Fu, P.~Wang, L.~Wu, R.~Bo, and X.~Li, ``Automating feature subspace
  exploration via multi-agent reinforcement learning,'' in \emph{Proceedings of
  the 25th ACM SIGKDD International Conference on Knowledge Discovery \& Data
  Mining}, 2019, pp. 207--215.

\bibitem{cruz2018improving}
F.~Cruz, S.~Magg, Y.~Nagai, and S.~Wermter, ``Improving interactive
  reinforcement learning: What makes a good teacher?'' \emph{Connection
  Science}, vol.~30, no.~3, pp. 306--325, 2018.

\bibitem{suay2011effect}
H.~B. Suay and S.~Chernova, ``Effect of human guidance and state space size on
  interactive reinforcement learning,'' in \emph{2011 Ro-Man}.\hskip 1em plus
  0.5em minus 0.4em\relax IEEE, 2011, pp. 1--6.

\bibitem{mnih2015human}
V.~Mnih, K.~Kavukcuoglu, D.~Silver, A.~A. Rusu, J.~Veness, M.~G. Bellemare,
  A.~Graves, M.~Riedmiller, A.~K. Fidjeland, G.~Ostrovski \emph{et~al.},
  ``Human-level control through deep reinforcement learning,'' \emph{Nature},
  vol. 518, no. 7540, pp. 529--533, 2015.

\bibitem{Dua:2019}
\BIBentryALTinterwordspacing
D.~Dua and C.~Graff, ``{UCI} machine learning repository,'' 2017. [Online].
  Available: \url{http://archive.ics.uci.edu/ml}
\BIBentrySTDinterwordspacing

\bibitem{van2000coil}
P.~Van Der~Putten and M.~van Someren, ``Coil challenge 2000: The insurance
  company case,'' Technical Report 2000--09, Leiden Institute of Advanced
  Computer Science~…, Tech. Rep., 2000.

\bibitem{yang1997comparative}
Y.~Yang and J.~O. Pedersen, ``A comparative study on feature selection in text
  categorization,'' in \emph{Icml}, vol.~97, no. 412-420.\hskip 1em plus 0.5em
  minus 0.4em\relax Nashville, TN, USA, 1997, p.~35.

\bibitem{granitto2006recursive}
P.~M. Granitto, C.~Furlanello, F.~Biasioli, and F.~Gasperi, ``Recursive feature
  elimination with random forest for ptr-ms analysis of agroindustrial
  products,'' \emph{Chemometrics and Intelligent Laboratory Systems}, vol.~83,
  no.~2, pp. 83--90, 2006.

\bibitem{peng2005feature}
H.~Peng, F.~Long, and C.~Ding, ``Feature selection based on mutual information
  criteria of max-dependency, max-relevance, and min-redundancy,'' \emph{IEEE
  Transactions on pattern analysis and machine intelligence}, vol.~27, no.~8,
  pp. 1226--1238, 2005.

\bibitem{kipf2016semi}
T.~N. Kipf and M.~Welling, ``Semi-supervised classification with graph
  convolutional networks,'' \emph{arXiv preprint arXiv:1609.02907}, 2016.

\bibitem{lin1992self}
L.-J. Lin, ``Self-improving reactive agents based on reinforcement learning,
  planning and teaching,'' \emph{Machine learning}, vol.~8, no. 3-4, pp.
  293--321, 1992.

\bibitem{guo2007semantic}
D.~Guo, H.~Xiong, V.~Atluri, and N.~Adam, ``Semantic feature selection for
  object discovery in high-resolution remote sensing imagery,'' in
  \emph{Pacific-Asia Conference on Knowledge Discovery and Data Mining}.\hskip
  1em plus 0.5em minus 0.4em\relax Springer, 2007, pp. 71--83.

\bibitem{kohavi1997wrappers}
R.~Kohavi, G.~H. John \emph{et~al.}, ``Wrappers for feature subset selection,''
  \emph{Artificial intelligence}, vol.~97, no. 1-2, pp. 273--324, 1997.

\bibitem{sutton2018reinforcement}
R.~S. Sutton and A.~G. Barto, \emph{Reinforcement learning: An
  introduction}.\hskip 1em plus 0.5em minus 0.4em\relax MIT press, 2018.

\bibitem{lin1991programming}
L.~J. Lin, ``Programming robots using reinforcement learning and teaching.'' in
  \emph{AAAI}, 1991, pp. 781--786.

\bibitem{wang2020incremental}
P.~Wang, K.~Liu, L.~Jiang, X.~Li, and Y.~Fu, ``Incremental mobile user
  profiling: Reinforcement learning with spatial knowledge graph for modeling
  event streams,'' in \emph{Proceedings of the 26th ACM SIGKDD International
  Conference on Knowledge Discovery \& Data Mining}, 2020, pp. 853--861.

\bibitem{knox2013teaching}
W.~B. Knox, P.~Stone, and C.~Breazeal, ``Teaching agents with human feedback: a
  demonstration of the tamer framework,'' in \emph{Proceedings of the companion
  publication of the 2013 international conference on Intelligent user
  interfaces companion}, 2013, pp. 65--66.

\bibitem{cederborg2015policy}
T.~Cederborg, I.~Grover, C.~L. Isbell, and A.~L. Thomaz, ``Policy shaping with
  human teachers,'' in \emph{Twenty-Fourth International Joint Conference on
  Artificial Intelligence}, 2015.

\bibitem{cruz2014improving}
F.~Cruz, S.~Magg, C.~Weber, and S.~Wermter, ``Improving reinforcement learning
  with interactive feedback and affordances,'' in \emph{4th International
  Conference on Development and Learning and on Epigenetic Robotics}.\hskip 1em
  plus 0.5em minus 0.4em\relax IEEE, 2014, pp. 165--170.

\bibitem{taylor2014reinforcement}
M.~E. Taylor, N.~Carboni, A.~Fachantidis, I.~Vlahavas, and L.~Torrey,
  ``Reinforcement learning agents providing advice in complex video games,''
  \emph{Connection Science}, vol.~26, no.~1, pp. 45--63, 2014.

\end{thebibliography}


\begin{thebibliography}{00}
\bibitem{b1} G. Eason, B. Noble, and I. N. Sneddon, ``On certain integrals of Lipschitz-Hankel type involving products of Bessel functions,'' Phil. Trans. Roy. Soc. London, vol. A247, pp. 529--551, April 1955.
\bibitem{b2} J. Clerk Maxwell, A Treatise on Electricity and Magnetism, 3rd ed., vol. 2. Oxford: Clarendon, 1892, pp.68--73.
\bibitem{b3} I. S. Jacobs and C. P. Bean, ``Fine particles, thin films and exchange anisotropy,'' in Magnetism, vol. III, G. T. Rado and H. Suhl, Eds. New York: Academic, 1963, pp. 271--350.
\bibitem{b4} K. Elissa, ``Title of paper if known,'' unpublished.
\bibitem{b5} R. Nicole, ``Title of paper with only first word capitalized,'' J. Name Stand. Abbrev., in press.
\bibitem{b6} Y. Yorozu, M. Hirano, K. Oka, and Y. Tagawa, ``Electron spectroscopy studies on magneto-optical media and plastic substrate interface,'' IEEE Transl. J. Magn. Japan, vol. 2, pp. 740--741, August 1987 [Digests 9th Annual Conf. Magnetics Japan, p. 301, 1982].
\bibitem{b7} M. Young, The Technical Writer's Handbook. Mill Valley, CA: University Science, 1989.
\end{thebibliography}

\end{document}